%% file: iclr2027_conference.tex
\documentclass{article} 
\usepackage{iclr2027_conference,times}

\input{math_commands.tex}

\usepackage{hyperref}
\usepackage{url}

\usepackage{hyperref}
\usepackage{algorithm}
\usepackage[noend]{algpseudocode}

\usepackage{amsmath}
\usepackage{amssymb}
\usepackage{mathtools}
\usepackage{amsthm}

\usepackage{physics}
\usepackage{multirow}
\usepackage{booktabs}       

\usepackage{arydshln}

\usepackage{color, colortbl}
\definecolor{LightCyan}{rgb}{0.88,1,1}
\definecolor{LightRed}{rgb}{1.0, 0.96, 0.96}

\usepackage[capitalize,noabbrev]{cleveref}

\usepackage{wrapfig}

\title{A Light Bilevel Refinement Aligns Self-Supervised Representations for Stronger Task-Specific Learning}

\author{Gustav Wagner Zakarias\textsuperscript{1,2},\quad Zheng-Hua Tan\textsuperscript{1,2}\\
    \textsuperscript{1}Aalborg University \quad\textsuperscript{2}Pioneer Centre for Artificial Intelligence, Denmark
    \\
    \texttt{[gwz,zt]@es.aau.dk}
}

\iclrfinalcopy 
\begin{document}

\maketitle

\begin{abstract}
Self-supervised pretraining learns representations that are broadly transferable across downstream tasks, yet direct fine-tuning can be suboptimal due to misalignment between self-supervised and downstream task objectives, potentially degrading pretrained features beneficial to the downstream task. The BiSSL framework addressed this by introducing a transitional training stage formulated as a bilevel optimization problem, in which the downstream task objective guides the self-supervised learning process in refining pretrained representations to better facilitate subsequent fine-tuning. However, BiSSL relies on conventional bilevel optimization solving techniques whose costly implicit hypergradient approximations render the method increasingly impractical for contemporary model architectures. To make it efficient and scalable, we introduce \emph{BiSSLight}, which combines M-FAC-based implicit gradient approximation with parameter-efficient fine-tuning via LoRA, enabling efficient application at larger scales that were previously impractical. Evaluation across multiple downstream tasks and contemporary model architectures shows that BiSSLight consistently improves downstream performance, with gains becoming more pronounced as model size increases despite stronger baselines. The method is highly computationally efficient, reducing computation time by more than a factor of ten compared to its predecessor on a  ViT-H backbone.
\end{abstract}

\section{Introduction}

Self-supervised learning (SSL) has become a dominant paradigm for learning transferable representations from large-scale unlabeled data~\citep{SimCLR, BYOL, MAE, ijepa, SSL_vision_survey_25}. SSL enables models to acquire general-purpose representations that can subsequently be adapted to downstream tasks through supervised fine-tuning, substantially alleviating the need for task-specific labeled data. However, the pretext task objectives and corresponding unlabeled data distributions encountered during SSL often differ considerately from those of the downstream task. Such misalignment can lead to inefficient adaptation when fine-tuned directly from the pretrained initialization, potentially causing degradation or overwriting of pretrained features that are beneficial to the downstream task~\citep{ft_distorts_pretrained_features, connect_later, connect_not_collapse, expl_forgetting_llms, ForgettingSurvey}.
\begin{figure}
    \centering
    \includegraphics[width=\linewidth]{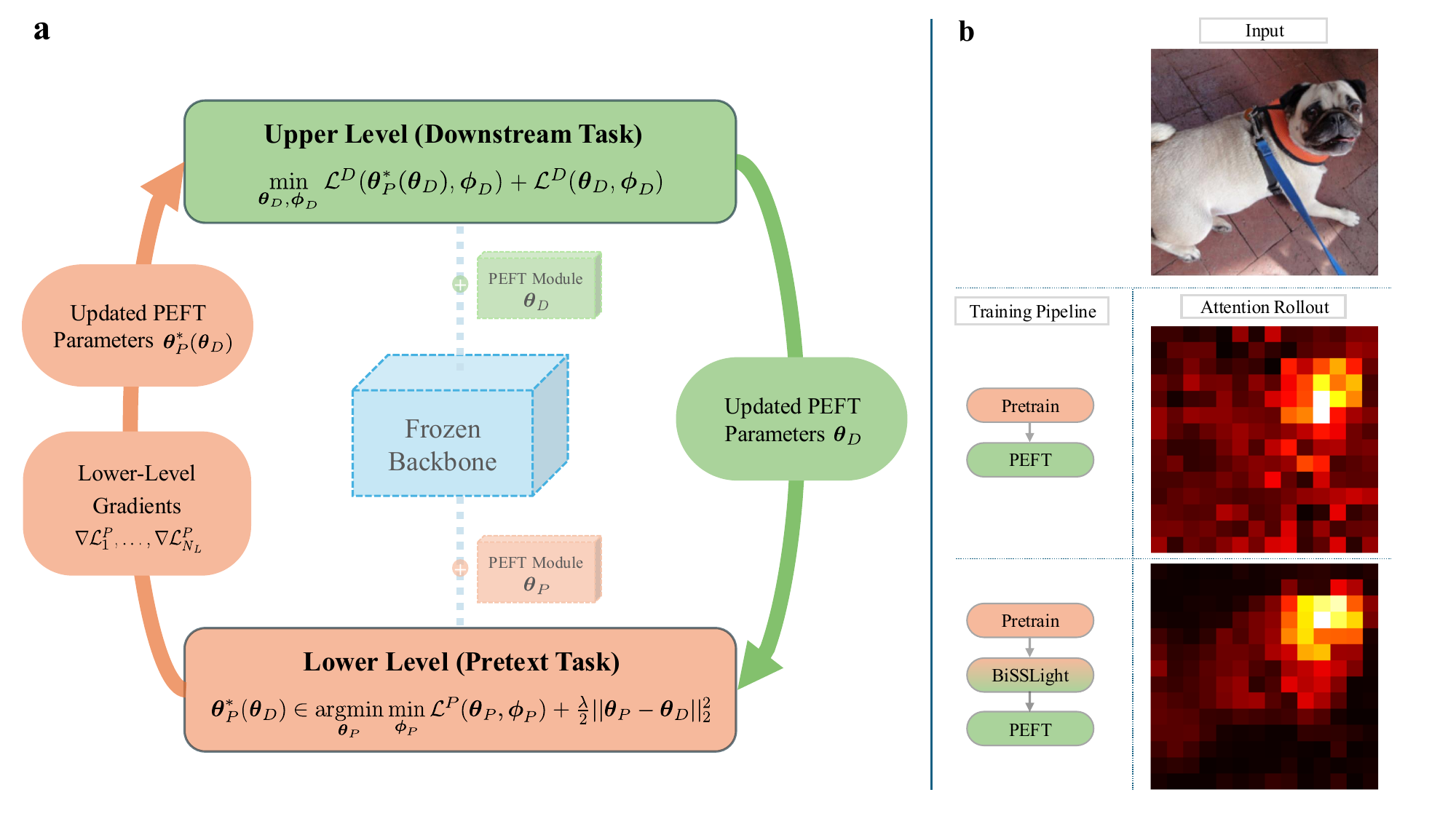}
    \caption{\textbf{a)} Overview of the BiSSLight training setup. Two distinct PEFT modules are attached to a shared frozen pretrained backbone and optimized alternately within the lower and upper-levels of the bilevel problem, respectively (see Section~\ref{subsec:revised_train_alg} for details). After training, the lower-level PEFT parameters \(\vb*{\theta}_P^\ast(\vb*{\theta}_D)\) are transferred to initialize downstream fine-tuning. \textbf{b)} Attention Rollout example comparing directly fine-tuning the pretrained model with fine-tuning the BiSSLight-trained model, suggesting that attention patterns are more concentrated on regions associated with the task-relevant object when applying BiSSLight. Further details and examples are provided in Section~\ref{sec:attn_rollout}.}
    \label{fig:bisslight_overview}
\end{figure}

The BiSSL framework~\citep{BiSSL} addresses this issue through the introduction of a transitional training stage that is applied to self-supervised pretrained backbones before downstream fine-tuning. This stage couples the self-supervised pretext and downstream objectives through a bilevel optimization problem~\citep{BLO_Survey}, allowing the downstream objective to guide the self-supervised learning toward refining the pretrained representations in a manner that better supports subsequent fine-tuning. As a result, BiSSL produces a backbone initialization that is better aligned with the downstream task, leading to consistently improved accuracy when fine-tuned across varying combinations of downstream and pretext tasks. However, these gains come at a substantial computational cost, largely due to the need to approximate implicit hypergradients during bilevel optimization, whose computational overhead grows unfavorably with scale. Consequently, BiSSL is poorly suited to the architectural scales commonly used in contemporary deep learning setups.

Meanwhile, modern downstream adaptation practice has increasingly shifted toward parameter-efficient fine-tuning (PEFT) methods~\citep{peft_vpt, peft_adaptformer, PEFT_PT_Survey} such as Low-Rank Adaptation (LoRA)~\citep{LORA}, reflecting a broader emphasis on scalability and computational efficiency when adapting large pretrained models. While BiSSL can in principle be combined with PEFT, its heavy bilevel optimization machinery largely negates the efficiency benefits that motivate PEFT in the first place, reducing its practical appeal.

In this work, we introduce \emph{BiSSLight}, an efficient and scalable variant of BiSSL that preserves its core objective of resolving misalignment between the pretext and downstream tasks. BiSSLight achieves this by employing the Matrix-Free Approximate Curvature (M-FAC) algorithm~\citep{M-FAC_ihvp_approx}, which we use for the first time in a bilevel optimization setting to approximate the upper-level implicit hypergradients without introducing additional hyperparameters. When combined with LoRA, this enables a training procedure that is substantially faster and more memory efficient than its predecessor. Importantly, BiSSLight retains the general applicability of the original framework, introducing no restrictions to the choice of pretext task, downstream task, or model architecture. Furthermore, it remains a drop-in intermediate training stage that can be applied to off-the-shelf pretrained backbones without requiring any modifications to either the pretraining or fine-tuning stages, offering straightforward integration into existing self-supervised transfer workflows. An overview of the method is shown in Figure~\ref{fig:bisslight_overview}.

\begin{figure}
    \centering
    \includegraphics[width=\linewidth]{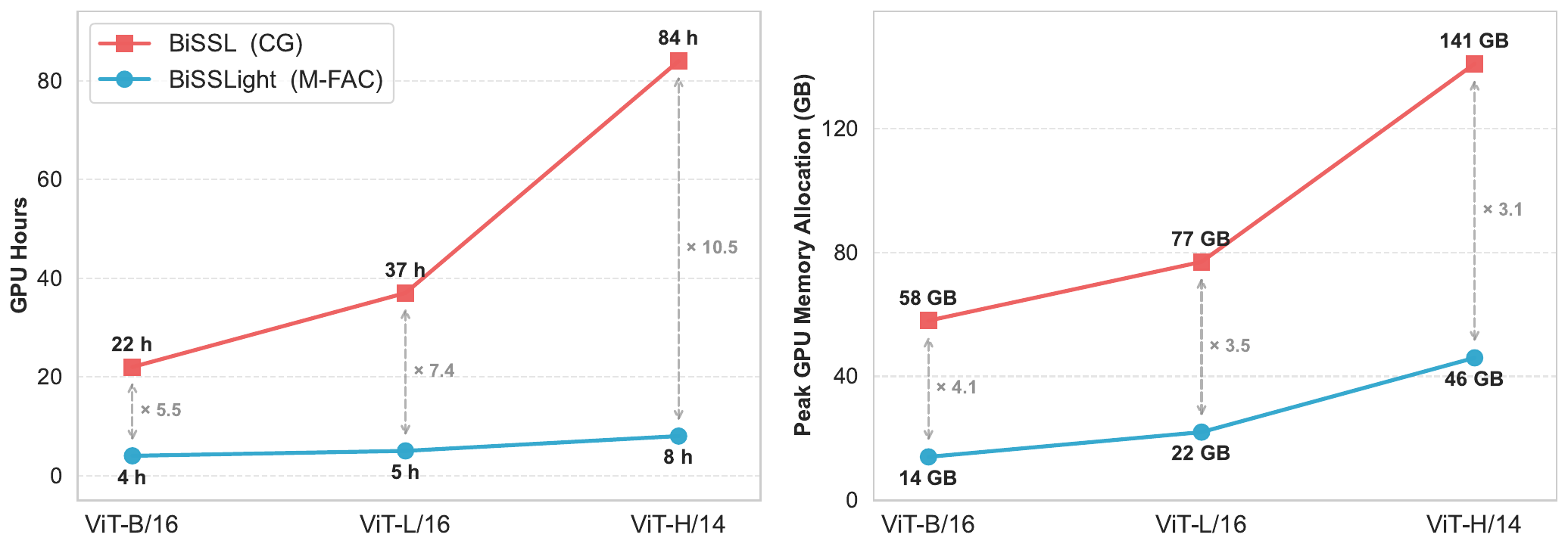}
    \caption{Computational scaling of BiSSL and BiSSLight with increasing backbone model size. Left: total computation time in GPU-hours. Right: peak GPU memory allocation. BiSSLight replaces the CG-based hypergradient approximation employed by BiSSL with M-FAC, which drastically reduces computational overhead. Notably, the relative runtime advantage becomes increasingly pronounced as model scale grows, and peak memory usage remains consistently several-fold lower.
    }
    \label{fig:compute_time_memory}
\end{figure}

We benchmark BiSSLight across a range of downstream tasks and backbone sizes. BiSSLight consistently improves downstream accuracy over directly fine-tuning pretrained models, with gains becoming more pronounced as model scale increases despite increasingly strong baselines. Importantly, as illustrated in Figure~\ref{fig:compute_time_memory}, these improvements are achieved with modest computational overhead, requiring only five GPU-hours to apply BiSSLight to a ViT-L/16 backbone. Additional experiments demonstrate robustness across a wide range of LoRA ranks and provide qualitative indications of more focused attention on task-relevant regions. Taken together, these results establish BiSSLight as an effective approach for improving downstream transfer in contemporary settings.

\section{Background}

\subsection{From Self-Supervised Pretraining To Downstream Fine-Tuning}
We first consider the standard pipeline of self-supervised pretraining followed by fine-tuning on a downstream task~\citep{SSL_vision_survey_25}. During self-supervised pretraining, we train a composite model of the form $h^P_{\boldsymbol{\phi}_P} \circ f_{\boldsymbol{\theta}}$ to solve a pretext task on an unlabeled dataset $\mathcal{D}^P = \{\vb{x}_i\}_{i=1}^{N_P}$, $\vb{x} \in \mathcal{X}$, where $f_{\boldsymbol{\theta}} : \mathcal{X} \mapsto \mathbb{R}^N$ is a feature-extracting backbone parameterized by $\boldsymbol{\theta}$, $h^P_{\boldsymbol{\phi}_P}:\mathbb{R}^N\mapsto\mathcal{Z}$ is a task-specific head with parameters $\boldsymbol{\phi}_P$ (e.g., the decoder in a masked autoencoder~\citep{MAE}) and \(\mathcal{Z}\) denotes the output space associated with the pretext task. Given the pretext task objective $\mathcal{L}^P(\boldsymbol{\theta}, \boldsymbol{\phi}_P;\mathcal{D}^P)$, self-supervised pretraining boils down to solving the optimization problem
\begin{equation}\label{eq:pretext_min_problem}
    \min_{\boldsymbol{\theta}, \boldsymbol{\phi}_P} \; \mathcal{L}^P(\boldsymbol{\theta}, \boldsymbol{\phi}_P;\mathcal{D}^P).
\end{equation}
After pretraining, the learned backbone parameters ${\boldsymbol{\theta}^\ast}$ are transferred to initialize training a downstream model in which the pretext task head is discarded and replaced by a downstream task-specific head. Given the labeled dataset $\mathcal{D}^D = \{(\vb{x}_j, \vb{y}_j)\}_{j=1}^{N_D}$ where $\vb{x} \in \mathcal{X}$, $\vb{y} \in \mathcal{Y}$, the model takes the form $h^D_{\boldsymbol{\phi}_D} \circ f_{\boldsymbol{\theta}}$, where $h^D_{\boldsymbol{\phi}_D}:\mathbb{R}^N \mapsto \mathcal{Y}$ is the downstream task head (e.g., a linear layer) with parameters $\boldsymbol{\phi}_D$. Given the downstream objective
$\mathcal{L}^D(\boldsymbol{\theta}, \boldsymbol{\phi}_D; \mathcal{D}^D)$, fine-tuning corresponds to solving
\begin{equation}\label{eq:ft_min_problem}
    \min_{\boldsymbol{\theta}, \boldsymbol{\phi}_D} \; \mathcal{L}^D(\boldsymbol{\theta}, \boldsymbol{\phi}_D;\mathcal{D}^D)\quad\text{s.t.}\quad \boldsymbol{\theta} \in \mathcal{N}({\vb*{\theta}^\ast}),
\end{equation}
where the notation reflects that fine-tuning is initialized from the pretrained parameter solution ${\boldsymbol{\theta}^\ast}$, with the fine-tuned backbone remaining in some neighborhood of the pretrained solution $\mathcal{N}(\boldsymbol{\theta}^\ast)$. For notational simplicity, we from now on omit the dataset specification from the training objective notation, e.g., $\mathcal{L}^P(\boldsymbol{\theta}, \boldsymbol{\phi}_P) := \mathcal{L}^P(\boldsymbol{\theta}, \boldsymbol{\phi}_P;\mathcal{D}^P)$.

\subsubsection{The Intermediate BiSSL Training Stage}\label{subsubsec:bissl_orig_intro}
Although fine-tuning adapts the pretrained representations to better comply with the downstream task, this adaptation may be inefficient when the representations induced by the pretext task are not fully aligned with the downstream task. To address this issue, BiSSL~\citep{BiSSL} introduces an intermediate alignment stage that is applied after pretraining and before standard downstream fine-tuning. This stage explicitly incorporates downstream supervision to guide the self-supervised learning process, steering the backbone toward downstream-aligned representations while preserving useful pretrained features. This provides a more favorable initialization, reducing the extent to which the subsequent fine-tuning procedure must reshape the learned representations and ultimately improving downstream performance. BiSSL formulates this transitional training stage as the following bilevel optimization problem:
\begin{align}
    \min_{\vb*{\theta}_D,\vb*{\phi}_D}&\quad \mathcal{L}^D(\vb*{\theta}_P^\ast(\vb*{\theta}_D),\vb*{\phi}_D) + \mathcal{L}^D(\vb*{\theta}_D,\vb*{\phi}_D)\label{eq:bissl_upper_level}\\
    \text{s.t.}&\quad \vb*{\theta}_P^\ast(\vb*{\theta}_D) \in \underset{\vb*{\theta}_P}{\text{argmin}}\,\min_{\vb*{\phi}_P}\, \mathcal{L}^P(\vb*{\theta}_P,\vb*{\phi}_P) + \frac{\lambda}{2}\norm{\vb*{\theta}_P - \vb*{\theta}_D}_2^2,\label{eq:bissl_lower_level}
\end{align}
Here, $\boldsymbol{\theta}_P$ and $\boldsymbol{\theta}_D$ denote two distinct backbone parameterizations, both initialized from the pretrained solution $\vb*{\theta}^\ast$. The regularization term in the lower-level objective in~\eqref{eq:bissl_lower_level} enforces proximity between the two backbones, constraining the lower-level optimization to remain close to the upper-level backbone while refining the learned representations. The hyperparameter $\lambda > 0$ controls the strength of this coupling. The upper-level objective in~\eqref{eq:bissl_upper_level} depends explicitly on the lower-level solution $\boldsymbol{\theta}_P^\ast(\boldsymbol{\theta}_D)$, which enables it to adjust its own backbone parameters $\boldsymbol{\theta}_D$ in a way that guides the lower-level optimization to benefit the downstream task. After solving the bilevel problem, the adapted backbone parameters $\boldsymbol{\theta}_P^\ast(\boldsymbol{\theta}_D)$ are used to initialize standard downstream fine-tuning as in~\eqref{eq:ft_min_problem}, replacing the original pretrained backbone $\boldsymbol{\theta}^\ast$. Unlike many common bilevel optimization setups, which often require considerable effort to design, implement, and tune optimization procedures for one or both levels~\citep{BLO_Survey}, BiSSL is comparatively straightforward to apply, as it directly reuses established pretext and downstream optimization procedures for the respective lower and upper-levels.

\paragraph{The Upper-Level Derivative}
The practical BiSSL training algorithm repeatedly alternates between $N_L$ lower-level and $N_U$ upper-level optimization steps for a total of $T$ iterations. While the lower-level gradient is comparatively straightforward to compute, the first term of the upper-level derivative with respect to the backbone parameters $\vb*{\theta}_D$ is more involved. Using the derivations provided in~\cite{BiSSL}, this derivative can be expressed as
\begin{align}\label{eq:bissl_upper_level_derivative}
    \frac{\text{d}\,\mathcal{L}^D(\vb*{\theta}^\ast_P(\vb*{\theta}_D),\vb*{\phi}_D)}{\text{d}\,\vb*{\theta}_D} =\lambda\qty[\nabla^2_{\vb*{\theta}}\mathcal{L}^P(\vb*{\theta},\vb*{\phi}_P)\vert_{\vb*{\theta}=\vb*{\theta}_P^\ast(\vb*{\theta}_D)} + \lambda I]^{-1}\nabla_{\vb{\theta}}\mathcal{L}^D(\vb*{\theta},\vb*{\phi}_D)\vert_{\vb*{\theta}=\vb*{\theta}_P^\ast(\vb*{\theta}_D)},
\end{align}
where $I$ is the identity matrix. A key practical challenge in evaluating this derivative lies in approximating the inverse of a high-dimensional Hessian matrix multiplied to a vector. The original BiSSL method adopts the conjugate gradient (CG) method~\citep{CG, CG_intro} to approximate this product, which is a canonical approach in bilevel optimization settings of this structure for computational tractability~\citep{BLO_Survey}. However, CG still requires calculating multiple sequential Hessian-vector products for each upper-level gradient estimate, which incurs substantial memory consumption and runtime overhead. As the training scale grows, this overhead becomes increasingly restrictive, making the original method progressively less suitable at larger scales.

\subsection{The M-FAC Algorithm}\label{subsec:mfac_intro}
The Fisher information matrix provides a surrogate for the Hessian under the assumption that a model’s output distribution matches its training data distribution~\citep{fisher_info_tutorial, woodfisher_ihvp_approx}. Under this equivalence, we can reasonably approximate the Hessian from a loss function $\mathcal{L}$ via the empirical Fisher information matrix, defined as $\hat{F} = \frac{1}{\vert\mathcal{D}\vert}\sum_{\vb{x} \in \mathcal{D}} \nabla_{\vb{\theta}} \mathcal{L}(\vb*{\theta};\vb{x})\nabla_{\vb{\theta}} \mathcal{L}(\vb*{\theta};\vb{x})^\top$, where $\mathcal{D}$ denotes a dataset and $\vb*{\theta}$ the trainable model parameters. Since $\hat{F}$ is a sum of rank-one matrices, we can compute its inverse-vector products efficiently using the Woodbury-Sherman-Morrison formula~\citep{woodbury-1950}. Given a collection of $m$ gradients $\nabla\mathcal{L}_1,\hdots,\nabla\mathcal{L}_m$, the Matrix-Free Approximate Curvature (M-FAC) algorithm~\citep{M-FAC_ihvp_approx} provides an efficient recursive equation of inverse empirical Fisher-vector products based on this formula, given by
\begin{align}\label{eq:mfac_recursion_general}
    \hat{F}^{-1}_i\vb{q} = \hat{F}_{i-1}^{-1}\vb{q} - \frac{\hat{F}_{i-1}^{-1}\nabla\mathcal{L}_i(\hat{F}_{i-1}^{-1}\nabla\mathcal{L}_i)^\top}{m + \nabla\mathcal{L}_i^\top\hat{F}_{i-1}^{-1}\nabla\mathcal{L}_i}\vb{q} =\gamma^{-1}\vb{q} - \sum_{j=1}^{i} \frac{\vb{v}_j^\top\vb{q}}{m+\nabla \mathcal{L}_j^\top\vb{v}_j}\vb{v}_j,
\end{align}
for $i=1,\hdots,m$, where $\vb{v}_i = \hat{F}_{i-1}^{-1}\nabla\mathcal{L}_i$ and $\vb{q}$ is an arbitrary query vector with initialization $\hat{F}_0^{-1}\vb{q}=\gamma^{-1}\vb{q}$ for a small $\gamma>0$. The vectors $\vb{v}_i$ are computed recursively using the expression above, and the final quantity $\hat{F}^{-1}_m \vb{q}$ serves as an approximation of $H^{-1}\vb{q}$. Notably, this recursion effectively approximates the inverse of $\hat{F} + \gamma I$, rather than $\hat{F}$, to ensure invertibility. This distinction plays an important role in the application of M-FAC within BiSSL, as discussed in Section~\ref{subsec:bissl_approx_ihvps}.

A key practical advantage of this formulation is that all vectors $\vb{v}_i$ and associated scalar denominators $m + \nabla\mathcal{L}_i\top\vb{v}_i$ depend only on the stored gradients and can therefore be computed once, independently of the query vector $\vb{q}$. Once these quantities are available, applying the inverse Fisher approximation to a new query vector requires only a small number of inner products and vector additions, as evident in~\eqref{eq:mfac_recursion_general}, making each evaluation of $\hat{F}^{-1}_m \vb{q}$ computationally inexpensive.

\paragraph{Related Work}
The related work discussion is deferred to Appendix~\ref{appsec:related_work}.

\section{BiSSLight: A Scalable Rewiring of BiSSL}\label{sec:bisslight}
We streamline BiSSL into \textit{BiSSLight} by integrating M-FAC for scalable and computationally efficient inverse Hessian-vector product approximation and LoRA for reducing memory usage.

\subsection{Approximating Inverse Hessian-Vector Products}\label{subsec:bissl_approx_ihvps}
The upper-level derivative in~\eqref{eq:bissl_upper_level_derivative} involves an inverse Hessian-vector product (IHVP), which constitutes the main computational bottleneck. To mitigate this, we aim to replace the conjugate gradient (CG)-based approximation with the M-FAC algorithm, covered in Section~\ref{subsec:mfac_intro}. The Hessian appearing in the upper-level gradient of~\eqref{eq:bissl_upper_level_derivative} is (in shorthand) $H= \nabla^2_{\vb*{\theta}_P}\mathcal{L}^P + \lambda I$. Recalling that M-FAC approximates $(H + \gamma I)^{-1}$, we can exploit the structure of $H$ by applying M-FAC only to the pretext task Hessian term $\nabla^2_{\vb*{\theta}_P}\mathcal{L}^P$ while setting $\gamma=\lambda$. The regularization term $\lambda I$ is thereby incorporated implicitly, which conveniently yields an approximation of exactly $H^{-1}$ in this setting. Since BiSSL is initialized from a pretrained model that performs the pretext task well, the empirical Fisher is expected to capture relevant local curvature information of the pretext task objective, making it a useful surrogate for the Hessian and thus rendering M-FAC well suited here.

\paragraph{Recycling Lower-Level Gradients} 
Unlike CG, M-FAC relies on a collection of gradients rather than repeated Hessian-vector products. This aligns naturally with the alternating BiSSL optimization procedure, as we can reuse the $N_L$ pretext-task minibatch gradients computed prior to each lower-level update for the M-FAC approximation, thereby avoiding computing any additional gradients. Although these gradients are evaluated at slightly different parameter values, the regularization term in~\eqref{eq:bissl_lower_level} keeps them close to the upper-level parameters, limiting drift during the $N_L$ updates and making their reuse of a reasonable foundation for the M-FAC approximation.

\paragraph{Block-Wise Approximation}
Since the M-FAC approximation is constructed from $N_L$ gradients, the contribution of the pretext task objective to the inverse Hessian approximation has rank at most $N_L$. In high-dimensional models, such a global low-rank approximation can be overly coarse. We therefore apply M-FAC in a block-wise manner, approximating inverse Hessian-vector products separately within each parameter block. This yields an $N_L$-rank approximation \textit{within each block}, resembling a block-diagonal curvature approximation commonly used in large-scale second-order optimization~\citep{woodfisher_ihvp_approx, kfac_ihvp_approx, shampoo}. Although cross-block interactions are ignored, the resulting approximation is distributed more evenly across the network.

We now derive the block-wise recursion used to compute the vectors $\vb{v}_i$ used in M-FAC based on~\eqref{eq:mfac_recursion_general}. Let $(\vb{q})_{\langle k \rangle}$, $k=1,\hdots,K$ denote a block partition of a vector $\vb{q}$. Given $N_L$ stored lower-level pretext task minibatch gradients $\nabla \mathcal{L}^P_1,\hdots,\nabla\mathcal{L}^P_{N_L}$, the recursion for block $k$ is
\begin{align}\label{eq:bissl_mfac_vi_calc}
    \qty(\vb{v}_i)_{\langle k\rangle} = \qty(\hat{F}^{-1}_{i-1}\nabla \mathcal{L}^P_i)_{\langle k\rangle} \approx \lambda^{-1}\qty(\nabla \mathcal{L}^P_i)_{\langle k\rangle} - \sum_{j=1}^{i-1} \qty(\vb{v}_j)_{\langle k\rangle}\frac{\qty(\vb{v}_j)_{\langle k\rangle}^\top\qty(\nabla \mathcal{L}^P_i)_{\langle k\rangle}}{N_L+ \underbrace{\qty(\vb{v}_j)_{\langle k\rangle}^\top \qty(\nabla \mathcal{L}^P_j)_{\langle k\rangle}}_{(d_j)_{\langle k \rangle}}},
\end{align}
for $i=1,\hdots, N_L$ where we define $(d_i)_{\langle k \rangle}:=\qty(\vb{v}_i)_{\langle k \rangle}^\top\qty(\nabla \mathcal{L}^P_i)_{\langle k \rangle}$ and also store these values in practice. Aside from storing $K\cdot N_L$ scalar denominators instead of $N_L$, a negligible overhead, this block-wise formulation incurs no additional storage overhead compared to a global approximation. The computational overhead is likewise negligible and can be parallelized across blocks due to the absence of cross-block dependencies, although we do not exploit this parallelism in our implementation. The recursion can either be executed online during lower-level optimization as new minibatch gradients become available, or after completion of the lower-level stage using the stored gradients. 

\subsubsection{The Updated Upper-Level Gradient Approximation}
After computing $(\vb{v}_1)_{\langle k\rangle},\ldots,(\vb{v}_{N_L})_{\langle k\rangle}$ and $(d_1)_{\langle k \rangle},\hdots,(d_{N_L})_{\langle k \rangle}$ via~\eqref{eq:bissl_mfac_vi_calc}, the resulting approximation to the $k$'th block of the first term of the upper-level derivative in~\eqref{eq:bissl_upper_level_derivative} is
\begin{align}\label{eq:upper_level_grad_approx_mfac}
    \qty(\frac{\text{d}\,\mathcal{L}^D(\vb*{\theta}^\ast_P(\vb*{\theta}_D),\vb*{\phi}_D)}{\text{d}\,\vb*{\theta}_D})_{\langle k\rangle} &=\lambda\qty(\qty[\nabla^2_{\vb*{\theta}}\mathcal{L}^P(\vb*{\theta},\vb*{\phi}_P)\vert_{\vb*{\theta}=\vb*{\theta}_P^\ast(\vb*{\theta}_D)} + \lambda I]^{-1}\nabla \mathcal{L}^D)_{\langle k\rangle}\\
    &\approx\lambda\qty(\hat{F}^{-1}_{N_L}\nabla \mathcal{L}^D)_{\langle k\rangle}\\
    &\approx \qty(\nabla \mathcal{L}^D)_{\langle k\rangle} - \lambda\sum_{j=1}^{N_L} (\vb{v}_j)_{\langle k\rangle}\frac{\qty(\vb{v}_j)_{\langle k\rangle}^\top\qty(\nabla \mathcal{L}^D)_{\langle k\rangle}}{N_L+ \qty(d_j)_{\langle k \rangle}}\label{eq:bissl_mfac_upper_grad_approx},
\end{align}
for each block $k=1,\hdots,K$ where $\nabla \mathcal{L}^D := \nabla_{\vb*{\theta}}\mathcal{L}^D(\vb*{\theta},\vb*{\phi}_D)\vert_{\vb*{\theta}=\vb*{\theta}^\ast_P(\vb*{\theta})}$. Beyond selecting the block partition, for which a layer-wise partitioning is often a sensible default, the approximation introduces no new hyperparameters requiring tuning, making it straightforward to adopt in place of the former CG-based approach. Since the computations in both~\eqref{eq:bissl_mfac_vi_calc} and~\eqref{eq:bissl_mfac_upper_grad_approx} involve only a small number of elementary vector operations, the associated M-FAC quantities and the inverse Hessian-vector product approximation can typically be computed on the CPU with negligible performance impact, thereby providing an additional opportunity to reduce GPU memory consumption. 

\subsection{Parameter-Efficient BiSSL with LoRA}
To further reduce the memory footprint, we freeze the pretrained backbone and apply LoRA modules with identical shapes at both the upper and lower-levels. Freezing the backbone eliminates the need to maintain separate full-scale models for each level, enabling BiSSL to operate on a single shared backbone with two lightweight sets of LoRA modules and thereby achieve a memory footprint comparable to that of standard PEFT. Furthermore, as LoRA restricts the lower-level optimization to a relatively low-dimensional parameter subspace, the M-FAC vectors in~\eqref{eq:bissl_mfac_vi_calc} are stored and manipulated only within this low-dimensional LoRA parameter space. This substantially reduces the memory and computational cost of the approximation, keeping the memory requirements of the approximation manageable even at large model scales.

From an optimization perspective, this modification does not alter the structure of the bilevel problem, as it is applied exactly as before, but with optimization carried out over the LoRA parameters instead. Concretely, we simply interpret $\vb*{\theta}_D$ and $\vb*{\theta}_P$ in~\eqref{eq:bissl_upper_level} and~\eqref{eq:bissl_lower_level} as the LoRA parameters rather than the full backbone weights, and afterwards transfer the fitted lower-level LoRA module $\vb*{\theta}_P^\ast(\vb*{\theta})$ as the initialization for subsequent standard PEFT.

\subsection{Revised Training Algorithm}\label{subsec:revised_train_alg}
The BiSSLight training procedure follows the same alternating bilevel structure as the original method, repeatedly alternating between lower-level and upper-level optimization phases. Before alternating back to the upper-level, the stored pretext task gradients are reused to compute the quantities required by the M-FAC recursion in~\eqref{eq:mfac_recursion_general}, which are then used for the upper-level gradient approximation as in~\eqref{eq:bissl_mfac_upper_grad_approx}. Figure~\ref{fig:bisslight_overview} provides a conceptual overview of the training procedure, while additional algorithmic details and pseudocode are provided in Appendix~\ref{appsec:bisslight_train_alg}.

\section{Experiments and Results}
This section benchmarks the impact of BiSSLight on modern high-capacity ViT backbone architectures~\citep{ViT}. The main experiments rely on the Masked Autoencoder (MAE) pretext task~\citep{MAE} and use the corresponding official ViT-B/16, ViT-L/16 and ViT-H/14 checkpoints, pretrained on ImageNet-1K~\citep{dset_imagenet} at a resolution of $224 \times 224$ pixels. MAE pretraining has demonstrated particularly strong performance when fine-tuned across a range of downstream tasks~\citep{mae_token_mergeing, peft_adaptformer, MAE_medical_appl, mae_plant_disease, mae_simpleclick, ExPLoRA}, making MAE-pretrained models a challenging and practically relevant setting for evaluating BiSSLight. Whereas the evaluation of the original BiSSL work was limited to ResNet-50 sized backbones~\citep{ResNet} using input resolutions of only $96\times96$ pixels, BiSSLight makes the framework practical under the model and input scales commonly adopted in modern computer vision research~\citep{MAE, dinov2, peft_for_vits}. Unless otherwise specified, we apply LoRA with rank $r=64$ to the query and value projection matrices in every attention layer, while keeping all pretrained backbone parameters frozen. Complete implementation and evaluation details are provided in Appendix~\ref{appsec:main_implementation_details}.

\begin{table}
    \caption{Comparison of classification accuracies obtained via directly applying PEFT on ViT-B/16, ViT-L/16 and ViT-H/14 backbones pretrained with the MAE pretext task and by applying BiSSLight prior to fine-tuning. Bold font indicates a statistically significant improvement in accuracy obtained with BiSSLight, occurring in the vast majority of the evaluated settings.}\label{tab:mae_bisslight_main_results}
      \centering
      \def\arraystretch{1.5}
      \resizebox{0.9\linewidth}{!}{
        \begin{tabular}{lcccccc}
        \toprule
          & Pets & DTD & VOC07 & Flowers & CUB200 & Aircrafts\\
        \midrule
        \textbf{ViT-B/16:}\\
        PEFT & \({91.7}\scriptstyle{\pm 0.3}\) & \({69.9}\scriptstyle{\pm 0.5}\) & \({80.9}\scriptstyle{\pm 0.3}\) & \({92.8}\scriptstyle{\pm 0.5}\) & \({79.8}\scriptstyle{\pm 0.5}\) & \({70.6}\scriptstyle{\pm 0.6}\) \\
        \rowcolor{LightCyan}
        BiSSLight+PEFT & \(\mathbf{92.1\scriptstyle{\pm 0.2}}\) & \({70.3\scriptstyle{\pm 0.4}}\) & \(\mathbf{81.7\scriptstyle{\pm 0.2}}\) & \(\mathbf{93.5\scriptstyle{\pm 0.3}}\) & \(\mathbf{81.2\scriptstyle{\pm 0.3}}\) & \(\mathbf{72.7\scriptstyle{\pm 0.3}}\) \\
        \cdashline{1-7}
        \textbf{ViT-L/16:}\\
        PEFT & \({94.4\scriptstyle{\pm 0.3}}\) & \({74.1\scriptstyle{\pm 0.6}}\) & \({85.5\scriptstyle{\pm 0.2}}\) & \({94.6\scriptstyle{\pm 0.4}}\) & \({84.4\scriptstyle{\pm 0.4}}\) & \({79.1\scriptstyle{\pm 0.4}}\) \\
        \rowcolor{LightCyan}
        BiSSLight+PEFT & \(\mathbf{94.7\scriptstyle{\pm 0.2}}\) & \(\mathbf{74.9\scriptstyle{\pm 0.4}}\) & \(\mathbf{85.9\scriptstyle{\pm 0.1}}\) & \(\mathbf{95.8\scriptstyle{\pm 0.2}}\) & \(\mathbf{85.6\scriptstyle{\pm 0.3}}\) & \(\mathbf{82.2\scriptstyle{\pm 0.3}}\) \\
        \cdashline{1-7}
        \textbf{ViT-H/14:}\\
        PEFT & \({95.2\scriptstyle{\pm 0.3}}\) & \({75.8\scriptstyle{\pm 0.5}}\) & \({86.5\scriptstyle{\pm 0.2}}\) & \({96.0\scriptstyle{\pm 0.3}}\) & \({85.9\scriptstyle{\pm 0.4}}\) & \({80.4\scriptstyle{\pm 1.0}}\) \\
        \rowcolor{LightCyan}
        BiSSLight+PEFT & \({95.5\scriptstyle{\pm 0.2}}\) & \(\mathbf{76.7\scriptstyle{\pm 0.2}}\) & \(\mathbf{87.3\scriptstyle{\pm 0.1}}\) & \(\mathbf{97.2\scriptstyle{\pm 0.1}}\) & \(\mathbf{87.2\scriptstyle{\pm 0.2}}\) & \(\mathbf{87.1\scriptstyle{\pm 0.3}}\) \\
        \bottomrule
      \end{tabular}
      }

\end{table}

\subsection{Downstream Task Performance}
We evaluate downstream performance on six image classification benchmarks spanning texture recognition (DTD)~\citep{dset_DTD}, multi-label object classification (VOC07)~\citep{VOC}, and fine-grained visual classification, including CUB200~\citep{CUB200}, Pets~\citep{dset_pets}, Flowers~\citep{dset_flowers}, and Aircrafts~\citep{dset_aircrafts}. 

Table~\ref{tab:mae_bisslight_main_results} reports classification accuracies obtained by direct fine-tuning of MAE-pretrained models and by applying BiSSLight prior to fine-tuning. Across the considered datasets, BiSSLight consistently improves the average accuracy while reducing performance variance across repeated runs. Moreover, in most cases, the observed accuracy gains are statistically significant.

A notable trend is that the benefits of BiSSLight increase with model scale. This is particularly evident on the aircrafts dataset, where the absolute gains grow from $+2.1$ percentage points for ViT-B to $+3.1$ for ViT-L and $+6.7$ for ViT-H, despite steadily improving baseline performance as model size increases. Similar behavior is observed on Oxford Flowers, where the gains increase from $+0.8$ percentage points for ViT-B to $+1.2$ for both ViT-L and ViT-H. Overall, these results suggest that the gains from BiSSLight are not exhausted by larger backbone capacities and may even be amplified when applied to stronger pretrained representations.

\paragraph{Different Pretext Task}
The original BiSSL framework was shown to provide downstream improvements across various pretext tasks. To verify that this property is preserved at larger model scales, we additionally evaluate BiSSLight using a ViT-H/14 backbone pretrained with the I-JEPA~\citep{ijepa} pretext task. Experimental details and results are provided in Appendix~\ref{appsubsec:ijepa}, showing that BiSSLight continues to provide consistent downstream performance gains.

\begin{figure}
    \centering
    \includegraphics[width=\linewidth]{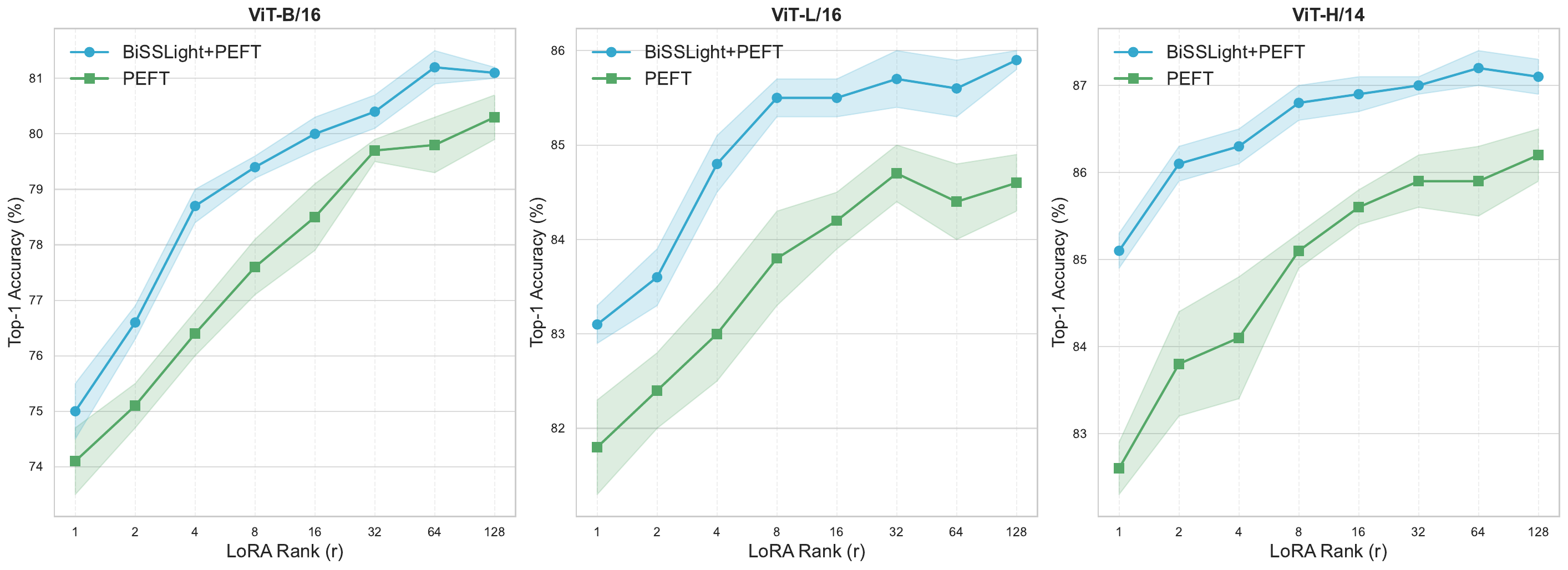}
    \caption{Test accuracies on the CUB200 dataset for varying LoRA ranks. Applying BiSSLight before PEFT consistently improves accuracy over directly applying PEFT across all evaluated ranks and backbone sizes, with gains generally becoming more pronounced at larger model scales.}
    \label{fig:lora_rank_var}
\end{figure}

\subsection{Performance Across LoRA Ranks}
To assess robustness across adaptation budgets, we vary the LoRA rank from $r=1$ to $r=128$ while keeping all other experimental settings identical to those described in Appendix~\ref{appsec:main_implementation_details}. We evaluate on the CUB200 dataset, which previously exhibited intermediate gains for BiSSLight compared to other benchmarks, making it a representative mid-range case. Figure~\ref{fig:lora_rank_var} shows that BiSSLight consistently improves downstream performance across all examined ranks and backbone scales. The relative improvements obtained by BiSSLight generally become larger as backbone scale increases, consistent with the trend observed in Table~\ref{tab:mae_bisslight_main_results}. Moreover, BiSSLight often matches or exceeds the performance of substantially higher-rank baseline models while using far fewer trainable parameters. On ViT-H for example, BiSSLight with only $r=2$ achieves performance comparable to the baseline with $r=128$.

\subsection{Comparing BiSSLight with BiSSL}\label{subsec:exp_compare_bissl}
To isolate the effect of the revised hypergradient approximation, BiSSL and BiSSLight are evaluated using identical backbone architectures, training schedules, hyperparameters, and LoRA configurations, as detailed in Appendix~\ref{appsec:main_implementation_details}. Consequently, the methods differ only in the hypergradient approximation procedure. Details of the comparison setup and the conjugate-gradient-based approximation used by BiSSL are provided in Appendix~\ref{appsubsec:exp_bissl_orig}.

\paragraph{Compute and Memory Costs}
Figure~\ref{fig:compute_time_memory} compares the computational cost of BiSSL and BiSSLight for increasing backbone sizes. BiSSLight substantially reduces both runtime and memory requirements. The runtime advantage becomes increasingly pronounced as model scale grows, improving from approximately \(5\times\) for ViT-B/16 to more than \(10\times\) for ViT-H/14. Peak memory consumption is reduced by roughly \(3\!-\!4\times\) across all evaluated architectures. Further analysis in Appendix~\ref{appsubsec:runtime_memory_bench} shows that the reported memory values are conservative upper-bound estimates, and that BiSSLight operates with a memory footprint broadly comparable to standard PEFT.

\paragraph{Accuracy and Representation Similarity}
Appendix~\ref{appsubsubsec:clsacc_rep_sim_bissl_orig} compares BiSSLight and BiSSL in terms of both downstream classification accuracy and representational similarity. BiSSLight achieves nearly identical downstream performance to the original BiSSL framework despite its substantially lower computational cost. The representational similarity analysis using Linear Centered Kernel Alignment (Linear CKA)~\citep{linear_cka} and Representational Alignment Analysis (RSA)~\citep{RSA} shows that BiSSLight learns highly similar representations to BiSSL. Moreover, both methods exhibit nearly identical representational similarity patterns relative to the pretrained backbone across datasets, further supporting the conclusion that they induce comparable representational changes during training. Taken together, these findings suggest that the efficiency gains have limited impact on the resulting representations and downstream performance.

\begin{figure}[t]
    \centering
    \includegraphics[width=\linewidth]{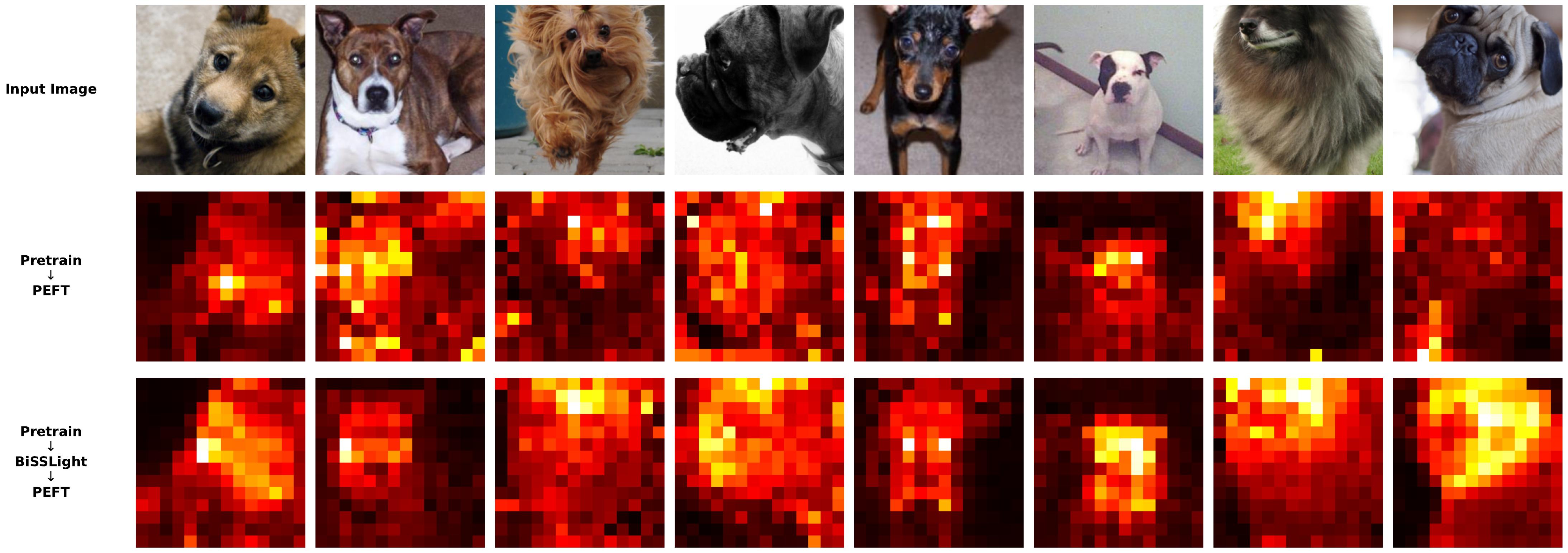}
    \caption{Attention rollout examples from the Pets validation set. The middle row corresponds to models directly fine-tuned from the MAE-pretrained checkpoint, while the bottom row shows models fine-tuned after the BiSSLight stage. BiSSLight produces attention patterns that are more concentrated on regions associated with the target object and less dispersed across the image.}
    \label{fig:attn_rollout_pets}
\end{figure}

\subsection{Attention Rollout Visualization}\label{sec:attn_rollout}
To qualitatively assess how BiSSLight influences the learned representations, we visualize attention-score maps, which are commonly used to study learned attention patterns in ViTs~\citep{emerging_properties_ssl_vit_attn_map, dinov2, dinov3, vit_needs_registers, ExPLoRA}. To obtain a compact model-wide view, we apply attention rollout~\citep{attn_rollout}, which aggregates attention across all transformer layers into a single relevance map, providing a holistic summary of attention propagation while remaining closely tied to the model's learned attention patterns. The visualizations are generated using a ViT-L/16 backbone, and results are shown for a randomly selected model acquired from one of the fine-tuning runs executed in conjunction with evaluating test accuracy (Described in Section~\ref{appsubsec:eval}). Results for representative examples from the Pets validation set are shown in Figure~\ref{fig:attn_rollout_pets}, and Appendix~\ref{appsubsec:attention_rollout} provides additional examples from the Flowers and CUB200 datasets together with further discussion. These datasets were selected because they contain visually well-defined foreground objects and classes whose distinguishing characteristics are readily interpretable by non-expert observers, making qualitative differences in attention allocation more accessible to evaluate. The BiSSLight-trained models frequently produce attention maps that appear more concentrated on semantically relevant image regions. In multiple cases, attention is distributed across a larger portion of the class-relevant target object and exhibits stronger contrast between object regions and surrounding background content, suggesting improved alignment between the learned representations and task-relevant visual features.

\section{Conclusion}
We presented BiSSLight, a scalable variant of BiSSL that enables efficient alignment of self-supervised representations with downstream objectives prior to fine-tuning. By combining M-FAC-based implicit gradient approximation with parameter-efficient fine-tuning (PEFT), BiSSLight significantly reduces computational and memory overhead while preserving the generality of the original framework, remaining agnostic to the choice of model architecture, pretext task, downstream task, and PEFT method. We showed that BiSSLight remains effective across a range of model sizes, consistently improving downstream performance across various tasks, with gains becoming more pronounced as models grow. Beyond the observed accuracy improvements, attention rollout visualizations are consistent with this alignment perspective, suggesting that BiSSLight tends to allocate attention more strongly to task-relevant image regions. Overall, BiSSLight extends the applicability of BiSSL to contemporary adaptation settings and provides a practical mechanism for improving the transition from self-supervised pretraining to downstream fine-tuning at scale. We hope this work underscores the importance of addressing pretraining-downstream misalignment prior to fine-tuning to fully leverage self-supervised pretrained models and encourages further exploration of algorithmic approaches to address this and related challenges.

\subsection*{AI use statement}
In this work, we used generative AI tools to suggest and edit code, provide feedback on the readability and clarity of portions of the manuscript, and assist in improving the presentation of figures by suggesting refinements to plotting code. AI-assisted writing support was limited to revising and refining manuscript text and occasionally drafting preliminary text based on author-provided notes. All such content was reviewed and substantially revised by the authors prior to inclusion in the final manuscript. We did not use generative AI tools to formulate the core methodology, design experiments, conduct the experimental evaluation, or determine the scientific conclusions of this work. The authors have reviewed all AI-assisted work, and take full responsibility for the final content of this work, including any material produced with the assistance of generative AI tools.

\subsection*{Reproducibility statement}
We make a dedicated effort to ensure the reproducibility of all results presented in this work. The proposed method is described in detail in Section~\ref{sec:bisslight} of the main paper, including the bilevel formulation, the M-FAC-based hypergradient approximation, and the incorporation of LoRA into the framework. Appendix~\ref{appsec:bisslight_train_alg} provides detailed pseudocode for the training algorithm, while Appendix~\ref{appsec:main_implementation_details} documents all implementation details required to reproduce the reported results. Additionally, we will publicly release the codebase together with the trained model checkpoints used in this work.\footnote{\url{https://github.com/42X4X01XZ/ICLR27_20859}}

\subsubsection*{Acknowledgments}
This project is supported by the Pioneer Centre for Artificial Intelligence, Denmark.\footnote{\url{https:\\www.aicentre.dk}} The authors would like to thank Lars Kai Hansen (DTU) for valuable feedback and insightful discussions that helped strengthen this work.

\bibliography{iclr2027_conference}
\bibliographystyle{iclr2027_conference}

\newpage
\appendix

\section{Related Work}\label{appsec:related_work}

\paragraph{Parameter-Efficient Fine-Tuning}
Parameter-efficient fine-tuning (PEFT) methods adapts large pretrained models by augmenting them with a relatively small number of trainable parameters while leaving the pretrained parameters fixed or by fitting only a small subset of the pretrained parameters~\citep{PEFT_PT_Survey}. Representative approaches include adapters, which insert trainable bottleneck modules throughout the network~\citep{peft_adapters_orig, peft_adaptformer}, prompt tuning, which prepends learnable prompts to the input sequence~\citep{peft_prompt_tuning_orig, peft_vpt}, prefix tuning, which injects trainable prefix vectors into the attention mechanism~\citep{peft_prefix_tuning, peft_prefix_tuning_vit}, BitFit, which fine-tunes only the bias-terms of the pretrained model~\citep{bitfit} and low-rank adaptation (LoRA), which introduces trainable low-rank matrix pairs whose product is added to selected pretrained weight matrices~\citep{LORA, peft_without_forgetting}. These methods reduce memory and computational cost compared to full fine-tuning and can also improve robustness to overfitting while maintaining strong downstream performance. Due to its simplicity and effectiveness, LoRA is widely used in practice and serves as the PEFT instantiation in this work, although BiSSLight can in principle be employed with other PEFT techniques as well.

\paragraph{Enhancing Downstream Adaptation of Pretrained Backbones}
Beyond BiSSL~\citep{BiSSL}, which is detailed in Section~\ref{subsubsec:bissl_orig_intro}, other works have similarly explored improving the downstream transferability of self-supervised representations prior to fine-tuning through an intermediate adaptation stage. ExPLoRA~\citep{ExPLoRA} adapts pretrained Vision Transformers through LoRA-based continued pretraining on downstream-domain data. Unlike ExPLoRA, which leverages downstream-domain data during the transitional stage and subsequently merges the adapted parameters back into the backbone before training a new set of adaptation parameters for the downstream task, BiSSLight directly incorporates the downstream objective into the optimization process and transfers the learned adaptation parameters directly to downstream PEFT. Other approaches instead conduct heuristic adaptation procedures intended to improve downstream adaptation. These include, among others, COIN~\citep{COIN}, which is tailored to contrastive pretraining through supervised contrastive learning on downstream data, NoisyTune~\citep{NoisyTune}, which perturbs the pretrained weights with backbone parameter-dependent noise, and RIFLE~\citep{RIFLE}, which repeatedly reinitializes the classification head. In contrast to the aforementioned approaches, BiSSLight do not assume that a particular adaptation procedure is generally beneficial across downstream tasks. Instead, the bilevel optimization problem allows the downstream task itself to guide the self-supervised learning process, resulting in an adaptation procedure tailored to the task at hand. 

\paragraph{Approximating Inverse Hessian Vector Products}
Among bilevel optimization methods that compute hypergradients via the implicit function theorem~\citep{dontchev2014implicit} such as BiSSL, a central challenge is the efficient approximation of the resulting inverse-Hessian vector products~\citep{BLO_Survey}. The most established approaches employ iterative procedures such as conjugate gradient (CG)~\citep{CG_intro, BLO_iMAML, BLOBeyondBackpropImp} and truncated Neumann series~\citep{neuman_ig_calc}, but their reliance on repeated Hessian-vector product computations still makes them increasingly expensive at scale. Simpler alternatives employ diagonal approximations to improve efficiency, at the cost of a coarser approximation of the Hessian~\citep{adahessian}. Between these extremes lies a family of methods that replace the Hessian with the empirical Fisher information matrix and exploit its structure to derive more efficient inverse-Hessian approximations~\citep{woodfisher_ihvp_approx}. One such method is Kronecker-factored Approximate Curvature (K-FAC)~\citep{kfac_ihvp_approx}, which exploits the layer-wise structure of deep neural networks to construct efficient approximations and has recently been explored for scalable hypergradient computation in bilevel optimization~\citep{kfac_blo}. However, K-FAC relies on architecture-specific structural assumptions and requires access to both activations and gradients, making the approximation more tightly coupled to the underlying model architecture and training setup. In contrast, Matrix-Free Approximate Curvature (M-FAC)~\citep{M-FAC_ihvp_approx} constructs the approximation solely from gradients, making it largely architecture- and PEFT-method agnostic and thereby helping BiSSLight preserve the plug-and-play nature of the original BiSSL framework.

\section{BiSSLight Training Algorithm}\label{appsec:bisslight_train_alg}
Algorithm~\ref{alg:bisslight} provides the complete pseudocode description of the resulting BiSSLight training procedure. The algorithm repeatedly alternates between solving the lower-level and upper-level optimization problems. After completing the lower-level optimization phase, the stored pretext gradients are used to compute the vectors and associated inner products required by the M-FAC recursion in~\eqref{eq:bissl_mfac_upper_grad_approx}. These quantities are stored in a collection $M$ and are subsequently reused when constructing the upper-level gradient approximation. An equivalent alternative is to compute the M-FAC quantities online during the lower-level optimization phase rather than after it has completed. Since both approaches perform the same operations, they have identical computational and memory complexity. We adopt the latter variant in the pseudocode, as it is also used in the provided PyTorch implementation.

To ensure a stable initialization of the bilevel optimization, we initialize the LoRA modules at both levels with identical parameter values, thereby avoiding an artificial mismatch between the upper- and lower-level parameterizations at the start of training. Throughout, we denote the pretrained (and frozen) backbone parameters by $\vb*{\psi}$, and the LoRA parameters associated with the lower- and upper-level objectives by $\vb*{\theta}_P$ and $\vb*{\theta}_D$, respectively.

\begin{algorithm}[H]
\caption{BiSSLight Training Algorithm}
\begin{algorithmic}[1]

\State \textbf{Input:} Initialized backbone and head parameters \(\vb*{\psi}\), \(\vb*{\phi}_P\), \(\vb*{\phi}_D\). Lower and upper-level training objectives \(\mathcal{L}^P\), \(\mathcal{L}^D\) and optimizers \(\text{opt}_P\), \(\text{opt}_D\). Regularization Weight \(\lambda\in\mathbb{R}_+\). Number of training stage alternations \(T\in\mathbb{N}\) with lower and upper-level steps \(N_L,N_U\in\mathbb{N}\). LoRA rank $r \in \mathbb{N}$. Block partition $1,\hdots, K$.
\item[]
\State Initialize rank-$r$ LoRA parameters $\vb*{\theta}_D$ on the backbone \(\vb*{\psi}\) and set $\vb*{\theta}_P \gets \vb*{\theta}_D$.
\item[]
\For{\(t=1,\hdots,T\)}
    \State Initialize $G \gets \emptyset$. \Comment{Lower-level gradient memory buffer}
    \For{\(i=1,\hdots,N_L\)} \Comment{Lower-level}
        \State Compute \(\nabla\mathcal{L}_i^P \gets \nabla_{\vb*{\theta}}\mathcal{L}^P\qty(\vb*{\theta}, \vb*{\phi}_P)\vert_{\vb*{\theta}=\vb*{\theta}_P}\) and append a copy to \(G\).
        \State Compute \(\vb{g}_{\vb*{\theta}_P} \gets \nabla\mathcal{L}_i^P  + \lambda \qty(\vb*{\theta}_P - \vb*{\theta}_D)\).
        \State Compute \(\vb{g}_{\vb*{\phi}_P} \gets \nabla_{\vb*{\phi}}\mathcal{L}^P(\vb*{\theta}_P,\vb*{\phi})\vert_{\vb*{\phi}=\vb*{\phi}_P}\).
        \State Update \(\vb*{\theta}_P \gets \text{opt}_P(\vb*{\theta}_P, \vb{g}_{\vb*{\theta}_P})\) and \(\vb*{\phi}_P \gets \text{opt}_P(\vb*{\phi}_P, \vb{g}_{\vb*{\phi}_P})\).
    \EndFor
    \item[]
    \State Compute $M=\qty[\qty((\vb{v}_i)_{\langle k\rangle},(d_i)_{\langle k\rangle})]_{i=1,\ldots,N_L,\;k=1,\ldots,K}$ using $G$ via~\eqref{eq:bissl_mfac_vi_calc}.

    \item[]
    \For{\(n=1,\hdots,N_U\)} \Comment{Upper-level}
        \State Compute \(\nabla\mathcal{L}^D \gets \nabla_{\vb*{\theta}}\mathcal{L}^D\qty(\vb*{\theta}, \vb*{\phi}_D)\vert_{\vb*{\theta}=\vb*{\theta}_P}\).
        \State Compute \(\vb{q}\) (assembled block-wise) using $M$ and $\nabla\mathcal{L}^D$ via~\eqref{eq:bissl_mfac_upper_grad_approx}.
        \State Compute \(\vb{g}_{\vb*{\theta}_D} \gets \vb{q} +  \nabla_{\vb*{\theta}}\mathcal{L}^D\qty(\vb*{\theta}, \vb*{\phi}_D)\vert_{\vb*{\theta}=\vb*{\theta}_D}\).
        \State Compute \(\vb{g}_{\vb*{\phi}_D} \gets \nabla_{\vb*{\phi}}\mathcal{L}^D(\vb*{\theta}_P,\vb*{\phi})\vert_{\vb*{\phi}=\vb*{\phi}_D} + \nabla_{\vb*{\phi}}\mathcal{L}^D(\vb*{\theta}_D,\vb*{\phi})\vert_{\vb*{\phi}=\vb*{\phi}_D}\).
        \State Update \(\vb*{\theta}_D \gets \text{opt}_D(\vb*{\theta}_D, \vb{g}_{\vb*{\theta}_D})\) and \(\vb*{\phi}_D \gets \text{opt}_D(\vb*{\phi}_D, \vb{g}_{\vb*{\phi}_D})\).
    \EndFor
    
\EndFor
\item[]
\State \textbf{Return:} Fitted LoRA Parameters \(\boldsymbol{\theta}_P\).

\end{algorithmic}\label{alg:bisslight}
\end{algorithm}

These modifications make BiSSLight practically compatible with modern large-scale training setups, with a memory footprint comparable to standard PEFT. The primary computational and memory cost now lies in the lower-level pretext optimization, as the upper-level gradient approximation contributes only minimal overhead. Because the lower-level optimization remains effective with relatively small batch sizes, the method can be deployed on hardware similar to that used for standard PEFT, as confirmed in the GPU memory requirement comparisons in Table~\ref{tab:memory_single_gpu} of Section~\ref{appsubsec:runtime_memory_bench}.

\section{Implementation Details of the Main Experimental Setup}\label{appsec:main_implementation_details}

\subsection{Self-Supervised Pretraining}
We exclusively use the official pretrained weights achieved via the Masked Autoencoder pretext task~\citep{MAE}. This includes both the encoder backbone and decoder head models, trained on the ImageNet-1k dataset~\citep{dset_imagenet} devoid of labels. We consider the three official ViT-B/16, ViT-L/16, and ViT-H/14 checkpoints pretrained at a resolution of $224 \times 224$. Full details of the MAE pretraining procedure are available in the original work~\citep{MAE}.

\subsection{Downstream Adaptation}\label{subsubsec:exp_downstream_ft_general}
For downstream fine-tuning, we apply standard data augmentations consisting of a random resized crop to $224 \times 224$ with a minimum crop ratio of $0.2$, followed by a random horizontal flip and normalization. We use the AdamW optimizer~\citep{AdamW} with weight decay $0.05$, $\beta_1=0.9$ and $\beta_2=0.999$, and train for $200$ epochs unless otherwise specified. Drop-path regularization~\citep{drop_path_reg, ViT} with rate $0.1$ is applied for ViT-B and rate $0.2$ for ViT-L and ViT-H.

We apply LoRA with rank $r=64$ to each respective query and value matrices in every attention layer, following common conventions~\citep{LORA, peft_without_forgetting, ExPLoRA}. A linear classification head is attached to the backbone output corresponding to the auxiliary class token, and only the LoRA parameters and this linear head are trained. We experimented with additionally tuning normalization layers, but observed no notable performance differences compared to freezing them. For simplicity and efficiency, we therefore keep these layers frozen. Similarly, we evaluated the use of dropout within LoRA modules, but found no measurable gains, and thus omit it, hypothesizing that the combination of drop-path regularization and weight decay provides sufficient regularization.

We adopt a cosine decaying learning rate scheduler with a linearly increasing warm-up~\citep{sgdr, MAE} over the first $5$ epochs. For models fine-tuned directly from pretrained backbones, we perform a grid search over the base learning rates $[10^{-4}, 3\cdot10^{-4}, 10^{-3}, 3\cdot10^{-3}, 10^{-2}, 3\cdot10^{-2}]$. For BiSSLight-trained backbones, we found that smaller learning rates are more suitable, likely due to the improved downstream alignment. Accordingly, we search over $[10^{-5}, 3\cdot10^{-5}, 10^{-4}, 3\cdot10^{-4}, 10^{-3}, 3\cdot10^{-3}]$. Besides this difference, BiSSLight-trained backbones are fine-tuned using the same setup as that used for directly fine-tuning the pretrained backbones.

\subsubsection{Evaluation}\label{appsubsec:eval}
After selecting the best learning rate based on the considered optimal combination of validation accuracy and loss, we perform \(8\) independent training runs with different random seeds. During each run, we retain the model checkpoint corresponding to the highest validation accuracy observed throughout training, and we evaluate validation accuracy after each epoch. Final results are reported as the mean and standard deviation of test accuracy across the \(8\) models.

To assess whether BiSSLight leads to statistically significant improvements over standard fine-tuning, we perform two-sample permutation tests comparing the classification accuracies obtained from the \(8\) runs of each method, and consider $p$-values below $0.05$ to imply a statistical significance between the average accuracies of the PEFT and BiSSLight+PEFT results. We refer to the original BiSSL work~\citep{BiSSL} for details, which uses the same class of test to asses statistical significance.

\paragraph{Downstream Datasets}
The main experiments evaluate six image classification benchmarks: Describable Textures Dataset (DTD)~\citep{dset_DTD}, PASCAL VOC 2007 (VOC07)~\citep{VOC}, CUB200~\citep{CUB200}, Oxford-IIIT Pet~\citep{dset_pets}, Oxford Flowers 102~\citep{dset_flowers}, and FGVC Aircraft~\citep{dset_aircrafts}. For datasets without official validation splits, we reuse the splits from~\cite{BiSSL}, which randomly allocate roughly $20\%$ of the training partition for validation while maintaining the same class balance. We use top-1 classification accuracy as the primary downstream metric for all datasets except VOC07, for which we report the standard 11-point mAP.

\subsection{BiSSLight}\label{appsubsec:exp_bisslight}
The experimental setup of BiSSLight does to a large extend follow the one of the original BiSSL work~\citep{BiSSL}, but we still describe all necessary implementation details here for completeness. We organize the description by first a remark on the LoRA configuration, then specifying level-specific settings, finally followed by details governing the overall bilevel training procedure.

\paragraph{LoRA Configuration}
Consistent with the fine-tuning setup described in Section~\ref{subsubsec:exp_downstream_ft_general}, we use the same LoRA rank during BiSSLight training to ensure compatibility and allow direct transfer of learned LoRA parameters. Unless otherwise stated, we hence apply LoRA with rank \(r=64\) to the respective query and value matrices in every attention layer and freeze all pretrained weights. Two identical sets of LoRA modules are instantiated for the lower- and upper-level objectives, respectively. Both sets are initialized with identical parameter values, after which they are optimized independently.

\paragraph{Lower-Level (Pretext Task)}
For the lower-level pretext task, we employ the same image augmentation pipeline as in the original MAE pretraining setup, which also matches the pipeline used during the downstream training stage: a random resized crop to \(224 \times 224\) with a minimum crop ratio of $0.2$, followed by a random horizontal flip and normalization. We similarly also train on the ImageNet-1k~\citep{dset_imagenet} devoid of labels. Although self-supervised pretraining typically relies on large batch sizes, preliminary testing showed that BiSSLight does not require large lower-level batches in order to perform well. We therefore use a batch size of \(256\), which also helps to keep the memory footprint further manageable.

We use the normalized pixel reconstruction loss from the original MAE implementation as well as a masking ratio of $0.75$. We apply a cosine decaying learning rate schedule with linear warm-up over \(10 \cdot N_L\) steps. The base learning rate for the LoRA backbone modules is set to \(0.01\). The pretrained MAE decoder head is fully unfrozen during lower-level training. While LoRA could also be applied to the decoder in highly memory-constrained settings, we found this unnecessary for our setup, as the decoder is per design comparatively small (\(\sim27\)M parameters). Since the decoder is initialized from pretrained weights, we use a smaller learning rate of \(10^{-4}\) for the decoder head parameters, which yields stabler training. We use the AdamW optimizer with \(\beta_1=0.9\), \(\beta_2=0.95\), and weight decay \(0.05\), following the original MAE configuration. Note that all lower-level settings are kept fixed across all downstream tasks.

\paragraph{Upper-Level (Downstream Task)}
The upper-level optimization configuration deliberately follows the downstream fine-tuning setup in Section~\ref{subsubsec:exp_downstream_ft_general} to a large extend. We use the same image augmentation pipeline, drop-path regularization, optimizer and cosine learning rate scheduler. The learning rate warm-up is performed over \(10 \cdot N_U\) steps and we use a batch size of $64$. For each dataset, we took the learning rate considered optimal for the pretext-pretrained-fine-tuned ViT-L model, and then obtained candidate learning rates by scaling this value by \(2/3\), \(1/2\), \(1/3 \), and, when necessary, by further reducing the learning rate within the same order of magnitude (e.g., \(3\cdot10^{-4}\),  \(2\cdot10^{-4}\), \(1\cdot10^{-4}\)).  The largest learning rate yielding stable optimization during continued BiSSLight training was ultimately selected.

Following the original BiSSL protocol, we initialize the downstream classification head using a brief linear probing phase prior to BiSSLight training. This improves training stability and convergence. The linear probing is performed for $20$ epochs using a fixed learning rate equal to that used for the upper-level BiSSLight optimization. For simplicity, we otherwise use exact same optimizer configuration as for the upper-level.

\paragraph{Overall Details}
We run BiSSLight as detailed in Algorithm~\ref{alg:bisslight} for a total of \(T=500\) alternations, with \(N_L=20\) lower-level and \(N_U=8\) upper-level optimization steps per alternation. For the lower-level optimization, this corresponds to just under two full pretext task epochs given the employed batch size. The regularization weight is fixed to \(\lambda=0.001\) across all experiments. These settings all follow the original BiSSL implementation. For the block-wise M-FAC approximation, we consider all parameters of each layer as a block, which in our case includes the collective LoRA modules assigned to the query and value matrices in each attention layer.

\subsection{LoRA Parameter Counts}
Table~\ref{tab:lora_param_count} reports the number of trainable LoRA parameters for each backbone and rank configuration. As expected, the parameter count scales linearly with the LoRA rank, while remaining a small fraction of the frozen backbone size even for the default $r=64$ used throughout this work.

\begin{table}
  \caption{Trainable LoRA parameter counts (in M) for the evaluated ViT backbones across different ranks. The default configuration used throughout the paper ($r=64$) corresponds to 2.75\%, 2.07\%, and 1.66\% of the parameters of the ViT-B/16, ViT-L/16, and ViT-H/14 backbones, respectively.}
  \label{tab:lora_param_count}
  \vspace{0.25cm}
  \centering
    \def\arraystretch{1.5}
  \resizebox{1.0\textwidth}{!}{
      \begin{tabular}{lcccccccc|c}
        \toprule
        \textbf{Parameters (M)} & $r=1$ & $r=2$ & $r=4$ & $r=8$ & $r=16$ & $r=32$ & $\mathbf{r=64}$ & $r=128$ & Full Backbone \\
        \midrule

        ViT-B/16 &\(0.04\) & \(0.07\) & \(0.15\) & \(0.30\) & \(0.59\) & \(1.18\) & \(\mathbf{2.36}\) & \(4.72\) & \(85.80\) \\
        ViT-L/16 & \(0.10\) & \(0.20\) & \(0.39\) & \(0.79\) & \(1.57\) & \(3.15\) & \(\mathbf{6.29}\) & \(12.58\) & \(303.30\) \\
        ViT-H/14 & \(0.16\) & \(0.33\) & \(0.66\) & \(1.31\) & \(2.62\) & \(5.24\) & \(\mathbf{10.49}\) & \(20.97\) & \(630.77\) \\
        \bottomrule
      \end{tabular}
  }
\end{table}

\section{Additional Results}

\subsection{I-JEPA Pretrained Backbone}\label{appsubsec:ijepa}

\subsubsection{Implementation Details}
We largely follow the experimental setup described in Appendix~\ref{appsec:main_implementation_details}, reporting only the modifications conducted for the I-JEPA~\citep{ijepa} setting below.

\paragraph{Self-Supervised Pretraining}
We use the official I-JEPA ViT-H/14 checkpoint at a resolution of $224\times224$, pretrained on ImageNet-1k~\citep{dset_imagenet}. We refer to the original work for the complete pretraining configuration.

\paragraph{Downstream Adaptation}
Since the pretrained I-JEPA model does not include a designated CLS token, we instead pass the average-pooled patch tokens from the final transformer layer through a batch normalization layer, followed by a linear layer. We use a weight decay of $0.0005$ and employ almost the same augmentation pipeline as of the main experiments in Section~\ref{subsubsec:exp_downstream_ft_general}, except that the minimum random crop ratio is increased from $0.2$ to $0.3$ to match the original I-JEPA setup.

\paragraph{BiSSLight} 
The lower-level training uses the same image augmentation pipeline employed during training of the original I-JEPA checkpoint. We initialize the teacher EMA schedule at $0.9995$ (instead of the original $0.996$), reflecting that training starts from a well-pretrained initialization and therefore does not require the teacher to track the student as closely during the early stages of optimization. We further use a weight decay of $0.0005$ and a LoRA learning rate of $0.005$. For the upper level, we also use a weight decay of $0.0005$ and set the minimum crop ratio to $0.3$ to remain consistent with the downstream fine-tuning setup.

\subsubsection{Classification Accuracies}
Table~\ref{tab:classificaiton_acc_bisslight_ijepa} reports downstream classification accuracies obtained using I-JEPA-pretrained models. BiSSLight consistently improves performance with statistical significance across all six benchmarks. Baseline accuracies are generally lower than those obtained with similarly sized MAE-pretrained models (see Table~\ref{tab:mae_bisslight_main_results}), which may partly reflect that I-JEPA was primarily designed with linear probing in mind rather than fine-tuning, as well as the fact that the I-JEPA checkpoint is trained for fewer epochs. Nevertheless, these results show that the benefits of BiSSLight extend beyond MAE pretraining and suggest that BiSSlight retains the ability of the original BiSSL framework to effectively adapt across pretext tasks.

\begin{table}
  \caption{Comparison of classification accuracies obtained via directly applying PEFT to a ViT-H/14 backbone pretrained with the I-JEPA pretext task and by applying BiSSLight prior to PEFT. Bold font indicates a statistically significant improvement in accuracy obtained with BiSSLight, occurring in all evaluated settings.}
  \label{tab:classificaiton_acc_bisslight_ijepa}
  \vspace{0.25cm}
  \centering
    \def\arraystretch{1.5}
  \resizebox{0.85\textwidth}{!}{
      \begin{tabular}{lcccccc}
        \toprule
        \textbf{I-JEPA} & Pets & DTD & VOC07 & Flowers & CUB200 & Aircrafts \\
        \midrule
        PEFT & \({94.6}\scriptstyle{\pm 0.1}\) & \({73.1}\scriptstyle{\pm 0.4}\) & \({82.4}\scriptstyle{\pm 0.2}\) & \({95.1}\scriptstyle{\pm 0.4}\) & \({86.4}\scriptstyle{\pm 0.3}\) & \({76.0}\scriptstyle{\pm 0.6}\) \\
        \rowcolor{LightCyan}
        BiSSLight+PEFT & \(\mathbf{{94.7}\scriptstyle{\pm 0.1}}\) & \(\mathbf{{74.0}\scriptstyle{\pm 0.6}}\) & \(\mathbf{{84.0}\scriptstyle{\pm 0.3}}\) & \(\mathbf{{96.1}\scriptstyle{\pm 0.3}}\) & \(\mathbf{{86.8}\scriptstyle{\pm 0.2}}\) & \(\mathbf{{77.7}\scriptstyle{\pm 0.4}}\) \\
        \bottomrule
      \end{tabular}
  }
\end{table}

\subsection{Comparing BiSSLight with BiSSL}\label{appsubsec:exp_bissl_orig}

\subsubsection{BiSSL Implementation Details}
For the experiments in Section~\ref{subsec:exp_compare_bissl} that compares BiSSLight with the original BiSSL framework~\citep{BiSSL}, we recreate the original optimization procedure by replacing M-FAC hypergradient approximation~\citep{M-FAC_ihvp_approx} with the conjugate-gradient (CG)~\citep{CG, CG_intro} approach used in BiSSL, while keeping all other experimental settings identical.

Following the original implementation, we perform five CG iterations per upper-level gradient estimate, corresponding to five Hessian-vector product evaluations per upper-level optimization step. We employ a damping value of $1$ (i.e., solving the system using $\lambda + 1$), consistent with standard CG practice, which we found sufficient for stable training. As a practical note, fused attention kernels are disabled during Hessian-vector product computations, as PyTorch fused attention kernels do not currently support the required second-order derivative calculations.

Apart from these modifications, the experimental setup is identical to that described in Appendix~\ref{appsec:main_implementation_details}. We refer the reader to the original BiSSL work for further details regarding the CG-based optimization procedure.

\subsubsection{Compute Time and Memory Benchmark Setup Details}\label{appsubsec:runtime_memory_bench}

BiSSL was executed on two NVIDIA A40 GPUs for ViT-B/16 and ViT-L/16, and four A40 GPUs for ViT-H/14 due to its substantially higher memory requirements. Although BiSSLight can be executed on a single A40 GPU for all evaluated backbones, we use the same GPU configurations as for BiSSL to ensure a fair comparison in Figure~\ref{fig:compute_time_memory}. Reported GPU-hours are computed as the total training runtime multiplied by the number of GPUs used.

Peak memory usage is measured on a per-epoch basis. Before each training epoch, the recorded CUDA memory statistics are reset, after which the peak memory allocation of each GPU is measured during the training phase of the epoch. The per-GPU peaks are summed and logged immediately after training each epoch, and the largest such value observed across the entire training run is reported. Since memory peaks on different GPUs may not occur simultaneously, the reported values should be regarded as conservative upper-bound estimates. However, the same measurement protocol is applied to both methods compared in Figure~\ref{fig:compute_time_memory}, making the reported values suitable for relative comparison.

The results are based on training on the CUB200 dataset, and since both methods execute a fixed number of optimization steps, runtime does not depend on the size of downstream dataset, making the reported numbers broadly representative. We remark that the reported memory and GPU-hour measurements should primarily be interpreted as relative comparisons rather than absolute reference values, as practical resource requirements depend on both the implementation and hardware.

\begin{table}
  \caption{Peak GPU memory requirements (in GB) for BiSSL, BiSSLight, and standard PEFT across different backbone scales and number of GPUs utilized, illustrating the practical memory footprint of BiSSLight relative to both BiSSL and conventional PEFT.}
  \label{tab:memory_single_gpu}
  \vspace{0.25cm}
  \centering
    \def\arraystretch{1.5}
  \resizebox{0.7\textwidth}{!}{
      \begin{tabular}{lc|ccc}
        \toprule
        & GPUs (ViT-B/L/H) & ViT-B/16 & ViT-L/16 & ViT-H/14 \\
        \midrule
        \rowcolor{LightRed}
        BiSSL & (2/2/4) & \(58\) GB & \(77\) GB & \(141\) GB \\
        \cdashline{1-5}
        \rowcolor{LightCyan}
        BiSSLight & (2/2/4) & \(14\) GB & \(22\) GB & \(46\) GB \\
        \rowcolor{LightCyan}
        BiSSLight & (1/1/1) & \(13\) GB & \(19\) GB & \(33\) GB\\
        \cdashline{1-5}
        PEFT, BS=$256$ & (1/1/2) & \(14\) GB & \(36\) GB & \(80\) GB\\
        PEFT, BS=$64$ & (1/1/1) & \(4\) GB & \(10\) GB & \(22\) GB\\
        \bottomrule
      \end{tabular}
  }
\end{table}

\paragraph{Single- versus Multi-GPU Memory Usage}
To provide additional context for the conservative memory measurements reported throughout the paper, we additionally evaluate BiSSLight on a single NVIDIA A40 GPU for each backbone and report the results in Table~\ref{tab:memory_single_gpu}. A corresponding comparison is not possible for BiSSL, as its memory requirements exceed the capacity of a single A40 GPU. The results show that the memory requirements reported throughout the paper are conservative, with peak memory usage often being lower when measured directly on a single GPU. This effect is particularly pronounced for ViT-H/14, where the measured peak allocation decreases from 46\,GB to 33\,GB. Part of this discrepancy stems from summing peak allocations across GPUs, while additional differences may arise from the inherent memory overheads associated with multi-GPU training.

Table~\ref{tab:memory_single_gpu} also reports the memory requirements of the subsequent parameter-efficient fine-tuning (PEFT) stage, with implementation details provided in Section~\ref{subsubsec:exp_downstream_ft_general}, measured on the same hardware. Under the batch-size configuration used throughout the paper, the PEFT stage requires more memory than BiSSLight. Reducing the PEFT batch size to match the BiSSLight upper-level optimization substantially lowers its memory footprint, indicating that the primary memory overhead of BiSSLight originates from the lower-level pretext optimization. Nevertheless, the resulting memory requirements remain broadly comparable, suggesting that BiSSLight can generally be deployed on hardware similar to that used for standard PEFT.

\begin{table}
  \caption{Comparison of classification accuracies obtained after applying either BiSSL or BiSSLight prior to fine-tuning using a ViT-B/16 backbone. BiSSLight generally reproduces the downstream performance gains of the original BiSSL framework across the evaluated datasets despite its substantially lower computational requirements.}
  \label{tab:classificaiton_acc_bissl_orig}
  \vspace{0.25cm}
  \centering
    \def\arraystretch{1.5}
  \resizebox{0.85\textwidth}{!}{
      \begin{tabular}{lcccccc}
        \toprule
        & Pets & DTD & VOC07 & Flowers & CUB200 & Aircrafts \\
        \midrule
        PEFT & \({91.7}\scriptstyle{\pm 0.3}\) & \({69.9}\scriptstyle{\pm 0.5}\) & \({80.9}\scriptstyle{\pm 0.3}\) & \({92.8}\scriptstyle{\pm 0.5}\) & \({79.8}\scriptstyle{\pm 0.5}\) & \({70.6}\scriptstyle{\pm 0.6}\) \\
        \rowcolor{LightCyan}
        BiSSLight+PEFT & \({92.1\scriptstyle{\pm 0.2}}\) & \({70.3\scriptstyle{\pm 0.4}}\) & \({81.7\scriptstyle{\pm 0.2}}\) & \({93.5\scriptstyle{\pm 0.3}}\) & \({81.2\scriptstyle{\pm 0.3}}\) & \({72.7\scriptstyle{\pm 0.3}}\) \\
        \rowcolor{LightRed}
        BiSSL+PEFT & \({92.0\scriptstyle{\pm 0.5}}\) & \({70.8\scriptstyle{\pm 0.4}}\) & \({81.9\scriptstyle{\pm 0.1}}\) & \({93.4\scriptstyle{\pm 0.4}}\) & \({81.2\scriptstyle{\pm 0.3}}\) & \({72.7\scriptstyle{\pm 0.3}}\) \\
        \bottomrule
      \end{tabular}
  }
\end{table}

\subsubsection{Similarity of Classification Accuracies and Representations}\label{appsubsubsec:clsacc_rep_sim_bissl_orig}
To verify that the efficiency improvements of BiSSLight do not come at the expense of downstream performance, Table~\ref{tab:classificaiton_acc_bissl_orig} compares BiSSLight against the original BiSSL framework on a ViT-B/16 backbone. BiSSLight achieves performance highly comparable to BiSSL across all datasets, differing by less than a fraction of a percentage point. Given the substantially higher computational demands of BiSSL, we restrict this comparison to a backbone size, which we consider sufficient to verify that the proposed efficiency improvements preserve downstream performance relative to the original framework.

\paragraph{Representational Similarity}
To additionally assess how similar the models learned via BiSSL and BiSSLight are, we compare the representations produced by the same models used in the downstream classification experiments for Table~\ref{tab:classificaiton_acc_bissl_orig}, namely the pretrained backbone, the BiSSL-trained backbone, and the BiSSLight-trained backbone for each downstream dataset. For each comparison, we compute two representational similarity metrics: Linear Centered Kernel Alignment (Linear CKA)~\citep{linear_cka} and Representational Similarity Analysis (RSA)~\citep{RSA, RSA_DNNs, RSA_DNN_Survey}. Linear CKA quantifies the alignment between two feature spaces based on their centered covariance structure, while remaining invariant to orthogonal transformations and isotropic scaling. For RSA, we first construct representational dissimilarity matrices (RDMs) using pairwise cosine distances between samples, and subsequently compute the Spearman correlation between the upper-triangular entries of the resulting RDMs. RSA therefore measures the extent to which two representations preserve the same pairwise relationships between samples. Further details regarding both metrics can be found in the original works.

\begin{figure}
    \centering
    \includegraphics[width=\linewidth]{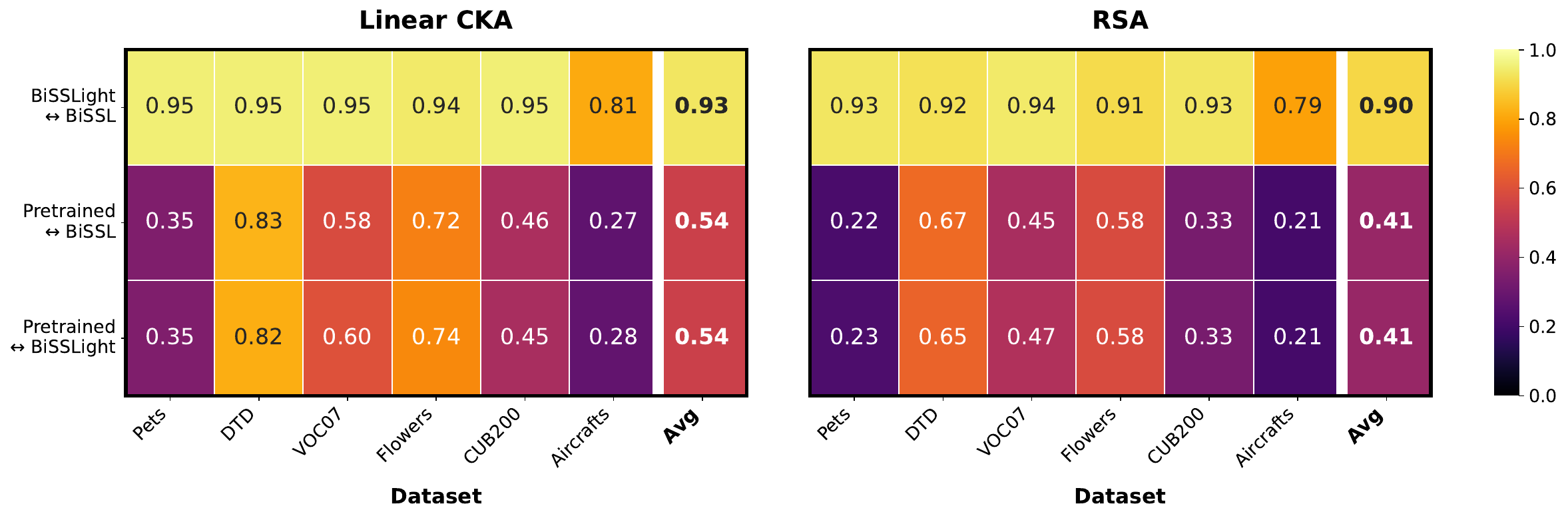}
    \caption{Dataset-specific representational similarity measured pair-wise using Linear CKA and RSA between combinations of the pretrained backbone, BiSSL-trained backbones and the BiSSLight-trained backbones. Each entry reports the representational similarity between two model variants evaluated on the indicated downstream dataset. Higher values indicate greater representational similarity. BiSSLight exhibits consistently high similarity to BiSSL across datasets while inducing comparable representational changes relative to the pretrained backbone.}
    \label{fig:repr_sim_per_dataset}
\end{figure}

Figure~\ref{fig:repr_sim_per_dataset} reports representational similarity scores between the pretrained backbone, BiSSL-trained models, and BiSSLight-trained models across all downstream datasets. Representations learned by BiSSL and BiSSLight exhibit consistently high similarity under both Linear CKA and RSA, with average scores of $0.93$ and $0.90$, respectively. While not equal to $1.0$, these values indicate a very high degree of representational similarity, particularly given the stochastic nature of the training and differences in the implementations of the two methods that may lead to minor deviations. Additionally, both methods exhibit substantially lower similarity to the pretrained backbone across all datasets, indicating that the bilevel training stage induces non-trivial representational changes. Lastly, the similarity scores between the pretrained backbone and each of the two methods are remarkably consistent across datasets, with average Linear CKA and RSA scores of $0.54$ and $0.41$, respectively, in both cases. While this alone does not imply that the resulting representations are similar, when considered together with the high similarity between BiSSL and BiSSLight, it provides additional evidence that suggest both methods induce comparable representational changes.

Taken together, these findings indicate that replacing the CG-based hypergradient approximation with M-FAC has limited effect on the resulting learned representations. This is consistent with the comparable downstream performance observed in Table~\ref{tab:classificaiton_acc_bissl_orig}, and implies that BiSSLight induces representational changes highly similar to those of the original framework while substantially reducing computational cost.

\begin{figure}[t]
    \centering
    \includegraphics[width=\linewidth]{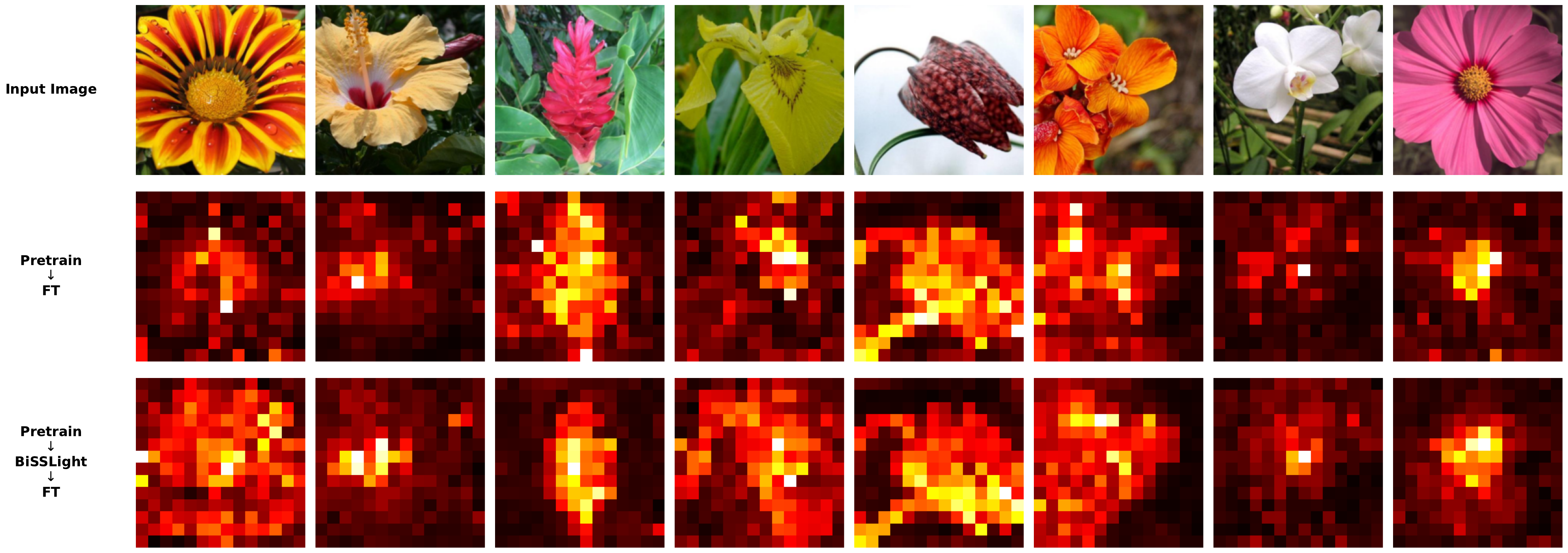}
    \caption{Attention rollout examples from the Flowers validation set. The middle row corresponds to models directly fine-tuned from the MAE-pretrained checkpoint, while the bottom row shows models fine-tuned after the BiSSLight stage. BiSSLight produces attention patterns that are more concentrated on regions associated with the target object and less dispersed across the image.}
    \label{fig:attn_rollout_flowers}
\end{figure}

\subsection{Attention Rollout Analysis}\label{appsubsec:attention_rollout}
\begin{figure}[t]
    \centering
    \includegraphics[width=\linewidth]{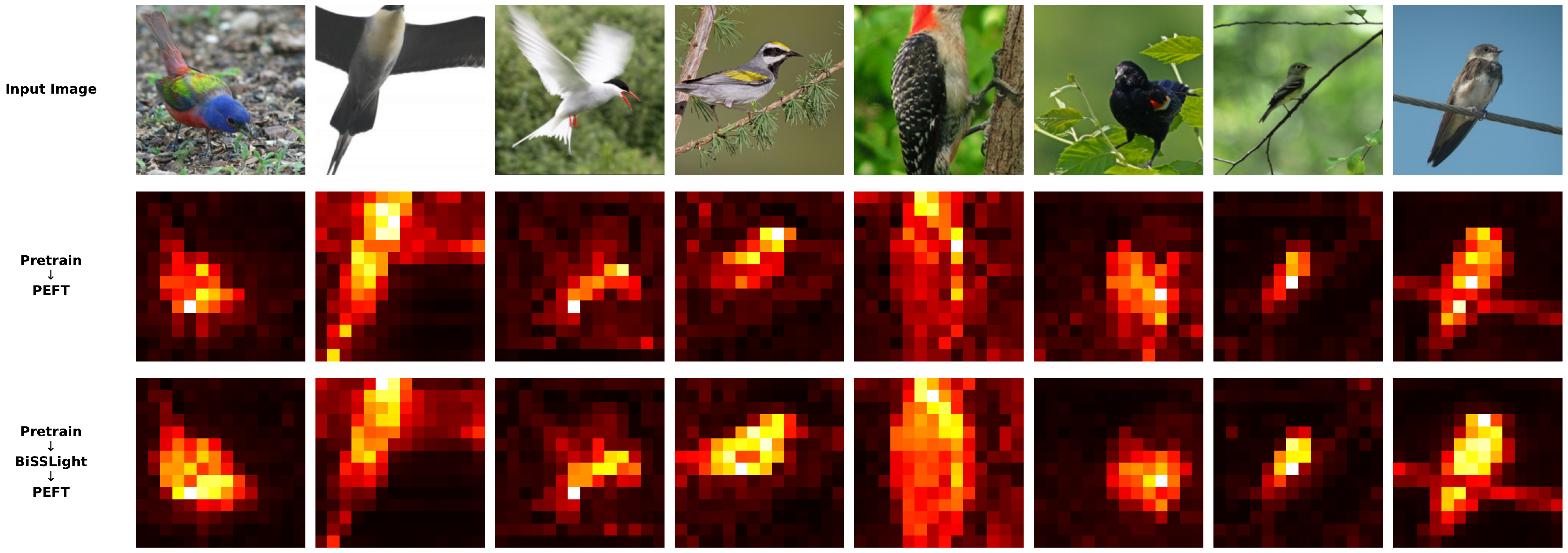}
    \caption{Attention rollout examples from the CUB200 validation set. The middle row corresponds to models directly fine-tuned from the MAE-pretrained checkpoint, while the bottom row shows models fine-tuned after the BiSSLight stage. BiSSLight produces attention patterns that are more concentrated on regions associated with the target object and less dispersed across the image.}
    \label{fig:attn_rollout_cub200}
\end{figure}

To complement the analysis in Section~\ref{sec:attn_rollout}, we include additional attention rollout maps for the Flowers and CUB200 datasets in Figures~\ref{fig:attn_rollout_flowers} and~\ref{fig:attn_rollout_cub200}, respectively. Similar to the examples shown for Pets in Figure~\ref{fig:attn_rollout_pets}, the BiSSLight-trained models frequently produce attention maps that appear more concentrated on semantically relevant image regions. In multiple cases, the highlighted patches are distributed across a larger portion of the target object and exhibit stronger contrast between object regions and surrounding background content.

Despite the relatively modest accuracy improvements observed on Pets in Table~\ref{tab:mae_bisslight_main_results}, the qualitative differences are often particularly pronounced for this dataset. One possible explanation is that Pets is already well aligned with the ImageNet pretraining distribution, allowing the baseline models to achieve strong performance using only a subset of the task-relevant visual information. Consequently, the downstream classification task may be relatively insensitive to further improvements in attention allocation, even when the resulting representations exhibit more object-focused behavior.

It should be noted that attention rollout provides only a highly condensed and approximate view of the model's internal attention patterns. Consequently, we use these visualizations primarily to assess whether systematic differences emerge between the attention maps produced by the two training procedures rather than to explain individual model predictions. Under this interpretation, the observed differences appear reasonably consistent across datasets.

\end{document}

%% file: math_commands.tex
\usepackage{amsmath,amsfonts,bm}

\def\eqref#1{equation~\ref{#1}}

\def\1{\bm{1}}

\def\vb{{\bm{b}}}

\DeclareMathAlphabet{\mathsfit}{\encodingdefault}{\sfdefault}{m}{sl}
\SetMathAlphabet{\mathsfit}{bold}{\encodingdefault}{\sfdefault}{bx}{n}

